\documentclass[letterpaper, 10 pt, conference]{ieeeconf}  

\usepackage{amssymb}
\usepackage{amsmath}
\usepackage{commath}
\usepackage{hyperref}
\usepackage{import}
\usepackage[pdf]{svg}

\IEEEoverridecommandlockouts                              

\usepackage{graphics} 
\usepackage{epsfig} 

\title{\LARGE \bf
Torque-Controlled Stepping-Strategy Push Recovery:\\
Design and Implementation on the iCub Humanoid Robot}

\author{Stefano Dafarra, Francesco Romano, Francesco Nori
\thanks{This work was supported by the FP7 EU projects Koroibot (No. 611909 ICT-2013.2.1 Cognitive Systems and Robotics)}%
\thanks{The authors are with the iCub Facility Department, Istituto Italiano di Tecnologia, 16163 Genova,
Italy (e-mail: name.surname@iit.it)}}

\DeclareMathOperator*{\minimize}{minimize}

\begin{document}
	
\maketitle
\thispagestyle{empty}
\pagestyle{empty}

\begin{abstract}
One of the challenges for the robotics community is to deploy robots which can reliably operate in real world scenarios together with humans.
A crucial requirement for legged robots is the capability to properly balance on their feet, rejecting external disturbances.
iCub is a state-of-the-art humanoid robot which has only recently started to balance on its feet.
While the current balancing controller has proved successful in various scenarios, 
it still misses the capability to properly react to strong pushes by taking steps.
This paper goes in this direction. It proposes and implements a control strategy based on the Capture Point concept \cite{Pratt2006}.
Instead of relying on position control, like most of Capture Point related approaches, the proposed strategy generates references for the momentum-based torque controller already implemented on the iCub, thus extending its capabilities to react to external disturbances
, while retaining the advantages of torque control when interacting with the environment.
Experiments in the Gazebo simulator and on the iCub humanoid robot validate the proposed strategy.

\end{abstract}

\section{INTRODUCTION}
\label{sec:intro}

An open robotics research problem is to endow robots with the capabilities to reliably operate in an unstructured environment, performing a variety of tasks at close contact with humans.
Indeed, robots must be able to move in the environment so as to accomplish their objectives.

 
The walking problem has been faced traditionally by resorting to simple models that approximate the predominant effects of the real systems.
One of the most popular models is the Linear Inverted Pendulum, whose 3D version has been introduced in \cite{Kajita2001}.
In particular, it has been applied to walking tasks \cite{Kajita2001,park1998biped}, or to predict when the robot is going to fall \cite{goswami2014direction} and,
also, within push recovery strategies.

Push recovery is a fundamental skill for a legged robot if it has to reliably operate in a real world scenario.
It embeds the ability to exploit the contacts with the environment in order to maintain stability in an erect position. 
The robotics community have come up with many efficient controllers to achieve this goal, such as \cite{Frontiers2015,Ott2011,koolen2016design}.
Sometimes, in case of severe disturbances, 
it may be necessary to take a step, in order to change the support configuration and reject the disturbance. 
This is what humans instinctively do. 

In literature this problem is tackled by resorting to simple models so as to obtain uncomplicated conditions to decide when to trigger the step and where the foot should be placed.
In the context of walking, the Linear Inverted Pendulum has been widely used \cite{stephens2007humanoid,morisawa2010combining,stephens2010pushforce}. 
The Capture Point framework \cite{Pratt2006,koolen2012capturability} has been used to stabilize walking patterns against external disturbances.
The strategy consists in controlling the dynamics of this particular point by generating a trajectory for the Zero Moment Point (ZMP) \cite{vukobratovic2004zero}.
This approach has been followed in \cite{englsberger2011bipedal,morisawa2012balance,krause2012stabilization} and implemented on position controlled robots.
In \cite{koolen2016design} the same approach has been used to stabilize the Atlas humanoid robot walking.
In \cite{ramos2014whole} authors introduce a Capture Point control ``task'' in the stack-of-task control formulation as a low priority objective.
The main motivation is to make the stack-of-task controller aware of the possibility to fall while reaching a desired target. 
Finally in \cite{pratt2012capturability} the Capture Point framework has been applied to the M2V2 force-controlled lower body humanoid robot.



This paper presents a control strategy based on the Instantaneous Capture Point concept to endow iCub with a stepping strategy.
The iCub humanoid robot is a state-of-the-art $53$ degrees of freedom robot \cite{metta2005robotcub}. 
Recently, a momentum-based whole-body torque controller \cite{Frontiers2015,nava16} has been synthesized, allowing the robot to balance on both two feet and one foot while performing complex movements with the limbs.
While this controller has proved successful in various balancing scenarios, it still misses the possibility to take a step in order to reject stronger disturbances, essential requirement for reliably operating together with humans.
This work proposes a surge in this direction, generating references to be fed to the momentum controller while retaining the advantages of the torque control. 
Experiments in simulation on Gazebo and on the real robot validate the proposed control architecture.


\section{BACKGROUND}
\label{sec:background}

This section introduces the model used in this work, together with the concept of the Capture Point.
In particular we describe a variation of the classic 3D Linear Inverted Pendulum model (LIP), namely the Linear Inverted Pendulum model with Finite-sized foot (FLIP) \cite{koolen2012capturability}.


\subsection{Notation} \label{sec:notation}
Throughout the paper the term 
 $\mathcal{I}$ denotes an inertial frame, with its $z$ axis pointing against the gravity, the $x$ axis points in front of the robot and with the origin placed on the ground level. We denote with $g$ the gravitational constant.
    
    Given a time function $f(t) \in \mathbb{R}^n$ its first and second order time derivatives are denoted as $\dot{f}(t)$ and $\ddot{f}(t)$ respectively.

\subsection{Linear Inverted Pendulum with Finite-sized foot model} \label{sec:flip}

\begin{figure}[t]
  \centering
    \def\svgwidth{0.8\columnwidth}
    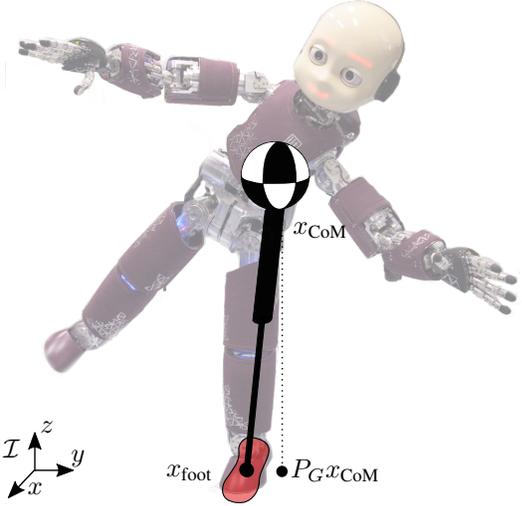
  \caption{3D Linear Inverted Pendulum with finite-sized foot model together with the full robot in background.}
  \label{fig:cubBalancing}
\end{figure}

The Linear Inverted Pendulum with Finite-sized foot model is an extension of the more classic LIP where a foot with finite dimension is introduced. 
The model approximates the lower body of a legged robot as an inverted pendulum with the point mass coinciding with the robot center of mass, connected by a massless rod to a foot in rigid contact with the ground as shown in Figure \ref{fig:cubBalancing}.
The foot possesses two degrees of freedom (DoFs), and thus can generate torques along the contact plane axes.
By assuming the center of mass to remain at a constant height during the robot motion, the model becomes linear.
It is thus possible to obtain the following equations of motion of the simplified model:
\begin{equation}
    \ddot{x}_{\text{CoM}} = \omega_0^2 ( P_G ~ {x}_{\text{CoM}} - x_{\text{CoP}})
\end{equation}
where $P_G$ projects on the ground plane the position of the center of mass ${x}_{\text{CoM}} \in \mathbb{R}^3$ (w.r.t. $\mathcal{I}$). $\omega_0$ is the natural frequency of the pendulum, i.e. $\omega_0 := \sqrt{g / z_0 }$, where $z_0$ is the initial CoM height.
Finally $x_{\text{CoP}}$ is the position of the center of pressure \cite{sardain2004forces} of the considered foot w.r.t $\mathcal{I}$.

\subsection{The Capture Point}

The Capture Point is defined as the point on the ground where the foot must be placed in order to stop the mass in the vertical upright position.
Mathematically, the Capture Point (CP), which can be found by considering the orbital energy of the pendulum \cite{Kajita2001}, 
takes the following expression:
\begin{equation}\label{eq:icp}
    x_\text{cp} = x_\text{CoM} + \frac{\dot{x}_\text{CoM}}{\omega_0}
\end{equation} 
and its dynamics is given by
\begin{equation}\label{eq:icp_dyn}
\dot{x}_\text{cp} = \omega_0(x_\text{cp}- x_\text{CoP}).
\end{equation}
The above equation is valid for the FLIP model. If the simple 3D LIP model is used, Eq. \eqref{eq:icp_dyn} becomes:
\begin{equation}\label{eq:icp_dyn_lip}
\dot{x}_\text{cp} = \omega_0(x_\text{cp}- x_\text{foot}).
\end{equation}

\subsection{Momentum-based whole-body torque control}
\label{sec:momentum}
The push recovery strategy proposed in this paper interfaces with the momentum-based whole-body torque control currently implemented on the iCub humanoid robot.
In this section we thus briefly describe its main peculiarities and we refer the reader to \cite{Frontiers2015,nava16} for additional details.

The momentum-based balancing controller is a hierarchical controller composed of two control objectives.
The first, and most priority objective, is the tracking of a desired robot momentum while the second is the stabilization of the zero dynamics.
Denoting with $H = [H_\text{lin}^\top, H_\text{ang}^\top]^\top \in \mathbb{R}^6$ the robot (linear and angular) momentum, its rate of change is obtained as $\dot{H} = \sum_{i = 1}^{n_c} {}^{\text{CoM}}X_i f_i + m \bar{g}$. Here $f_i \in \mathbb{R}^6 = [F_i^\top, \mu_i\top]^\top$ is the $i$-th of the $n_c$ contact wrenches composed of the 3D force and moment, ${}^\text{CoM}X_i \in \mathbb{R}^{6 \times 6}$ is the matrix transforming the corresponding wrench from the application frame to a frame attached to the center of mass with the same orientation of the inertial frame $\mathcal{I}$, $m$ is the robot total mass and $\bar{g} \in \mathbb{R}^6$ is the 6D gravity acceleration vector.
By assuming as virtual control inputs the contact wrenches $f = [f_1^\top, \cdots, f_{n_c}^\top]^\top$, it is possible to control the robot momentum by solving the following minimization problem:
\begin{equation}
    \label{eq:mom_min}
    \begin{aligned}
        \minimize_f ~& \norm{\dot{H} - \dot{H}^d}^2 \\
        \text{s.t.}~ & A f \leq b
    \end{aligned}
\end{equation}
where the inequality constraint $A f \leq b$ represents friction cone, center of pressure and other constraints on the wrenches.
The desired momentum rate of change is obtained by mean of a PI control law plus a feed-forward action \cite{nava16}. 

The second objective is responsible for constraining the joint variables and avoid internal divergent behaviors.
As before, we can specify a minimization problem also for this second task, i.e.
\begin{IEEEeqnarray}{rCl}
    \label{eq:zero_stab_min}
            \minimize_\tau&~ & \norm{\tau - \psi}^2  \IEEEyessubnumber \label{eq:zero_stab_min_cost}\\
            \text{s.t.}&& M(q)\dot{\nu} + h(q, \nu) - J^\top f = S \tau \IEEEyessubnumber \label{eq:zero_stab_min_dyn}\\
                       && J \dot{\nu} + \dot{J} \nu = 0  \IEEEyessubnumber \label{eq:zero_stab_min_constr}\\
                       && \psi := h_j(q, 0) - J^{(j),\top} f  \notag \\
                       && \quad\quad - K_p^j(q_j - q_j^r) - K_d^j \dot{q}_j \IEEEyessubnumber \label{eq:zero_stab_min_post}\\
                       && \norm{\dot{H} - \dot{H}^d}^2 = \text{solution of }\eqref{eq:mom_min}\IEEEyessubnumber \label{eq:zero_stab_min_hier}.
\end{IEEEeqnarray}
Eq.\eqref{eq:zero_stab_min_dyn} describes the free-floating dynamics of the mechanical system, with $q = (q_b, q_j)$, $q_b \in \mathbb{R}^3 \times SO(3)$ denotes the configuration of the floating base w.r.t. $\mathcal{I}$, $q_j \in \mathbb{R}^n$ denotes the configuration of the internal $n$ DoFs, $\nu$ is the velocity of the system, $J = [J_1^\top, \cdots, J_{n_c}^\top]$ is the stack of the contact Jacobians and $S \in \mathbb{R}^{n + 6 \times n}$ is a selector matrix describing the underactuation pattern.
Eq.\eqref{eq:zero_stab_min_constr} is the constraint equation describing the kinematic constraints associated with the contacts.
Eq.\eqref{eq:zero_stab_min_post}, which resembles a PD plus gravity and contact wrenches compensation, plays the role of a desired joint torque reference where $h_j$ and  $J^{(j)}$ denotes the joint space bias term and Jacobian respectively.
Finally Eq.\eqref{eq:zero_stab_min_hier} is the hierarchical constraint, i.e. it prevents  the solution of this second problem from changing the optimum of Eq.\eqref{eq:mom_min}.

Equations \eqref{eq:zero_stab_min_dyn} and \eqref{eq:zero_stab_min_constr} together describes the dynamics of the constrained dynamical system.
It is worth noting that when the constraint set changes, e.g. when the robot goes from two feet to one foot or vice versa, the constrained dynamics changes.
The overall system is thus a hybrid system and the discrete state transitions should be handled accordingly.
In the current implementation particular care has been taken in mitigating torque discontinuities happening during the transition.

%

Summarizing, from a functional point of view, the momentum-based controller takes as references a desired momentum trajectory, i.e. $\dot{H}^r, H^r$, a desired joint configuration $q_j^r$ and the set of contact constraints. The generated torques are then applied directly as references for the low-level torque control.

\section{The Stepping Strategy}
\label{sec:contribution}
\begin{figure*}[t]
	\centering
	\def\svgwidth{\textwidth}
	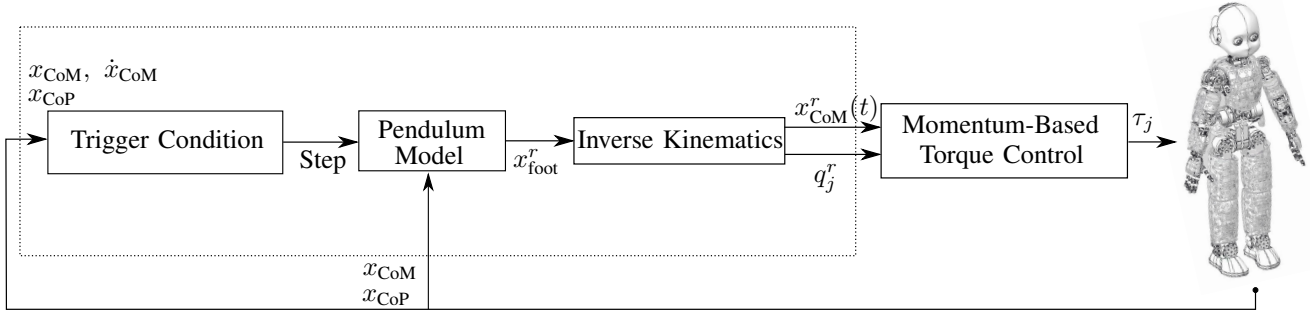
	\caption{Schema of the proposed Push Recovery Strategy.}
	\label{fig:strategy}
\end{figure*}

This section describes the control architecture necessary to implement the Stepping strategy on the iCub humanoid robot.
In particular, the proposed strategy is composed of three main phases:
 \begin{enumerate}
 	\item Step Trigger,
 	\item Foot Placement,
 	\item Reference Generation.
 \end{enumerate}
Peculiar to this implementation, and differently from most of the state-of-the-art approaches, is the fact that the resulting trajectories are fed to the momentum-based torque control currently implemented on the iCub robot (see Section \ref{sec:momentum}).

Figure~\ref{fig:strategy} depicts a schematic of the control architecture.

%
%
%
 
\subsection{Step Trigger}
\label{sec:trigger}
The first element composing the stepping strategy is the so called ``Step Trigger''. 
This component is responsible to detect when the internal torques are no longer sufficient to balance the robot and instead a change in the support configuration must be made, i.e. the robot must take a step.
 
In particular, we can define a ``do not step'' condition, i.e. a condition such that as long as it is satisfied there is no need to step, namely:
\begin{equation*} 
    \norm{ x_\text{cp} - x_\text{foot}} \leq \norm{x_\text{edge} - x_\text{foot}} 
\end{equation*}
where $x_\text{edge} \in \mathbb{R}^3$ basically defines the intersection on the foot contour of the line passing through $x_\text{cp}$ and $x_\text{foot}$. In other words, the step is triggered when the Capture Point exits the support polygon. 
This kind of trigger is derived from Eq.\eqref{eq:icp_dyn}. When the CP exits the foot boundary, the CoP cannot follow it, thus leaving the Capture Point to exponentially diverge from the foot.

%

%

\subsection{Foot Placement}\label{sec:foot_placement}
This second element is the core part of the control architecture, as it is responsible for choosing where the foot should be placed so as to avoid the robot falling. Indeed, the feet positions are not planned in advance and instead they should be computed online.

We start by considering for simplicity the dynamics of the Capture Point in the LIP model, i.e. Eq. \eqref{eq:icp_dyn_lip}.
The solution to the differential equation, with initial conditions in $t = 0$ is given by $x_\text{cp}(t) = e^{\omega_0 t}\left(x_\text{cp}(0)- {x}_\text{foot}\right) + {x}_\text{foot}$.
This equation describes the time evolution of the Capture Point, given the initial condition and considering that $x_\text{foot}$ remains constant.
In order to reject the disturbance, the foot is placed taking into account the Capture Point dynamic.
Hypothesizing that the time needed to perform a step is $t_\text{step}$ we want to position the foot in
\begin{equation}
    \label{eq:capture_point_predicted_lipm}
    x_\text{cp}(t_\text{step}) = e^{\omega_0 t_\text{step}} \left(x_\text{cp}(0)- {x}_\text{foot}\right)+ {x}_\text{foot}.
\end{equation}

If the FLIP model is used instead, the dynamic equation to be solved is the one in Eq. \eqref{eq:icp_dyn}.
Differently from before, this equation is not trivially integrable as the CoP is a function of time itself.
A classic approach, used for example in \cite{koolen2012capturability}, is to use a constant ``equivalent CoP''. Furthermore, a desired Capture Point trajectory is planned, and tracked by means of controlling directly the CoP itself.

Instead of keeping the desired foot location constant and tracking the desired Capture Point, we decide to continuously update the desired foot location by repeatedly initializing the solution at the current state of the robot.
At a generic time instant $0 \leq \bar{t} \leq t_\text{step}$ we integrate Eq. \eqref{eq:icp_dyn} with the further assumption that $x_\text{CoP}$ remains constant $\forall t : \bar{t} \leq t \leq t_\text{step}$.
The solution to this differential equation computed in $t = t_\text{step}$ is given by the following equation
\begin{equation}\label{eq:icp_estimation}
    x_\text{cp}(t_\text{step}) = e^{\omega_0 (t_\text{step} - \bar{t})}\left(x_\text{cp}(\bar{t})- {x}_\text{CoP}(\bar{t}) \right)+ {x}_\text{CoP}(\bar{t}).
\end{equation}
The adoption of the CoP as a feedback variable, allows to have a hint about the capability of the underneath balancing controller in controlling the CoM dynamics and, consequently, the Capture Point dynamics. Its effects (CoP and CP motion) are taken into account inside Eq. \ref{eq:icp_estimation}, so that it may be possible to perform a step smaller than what predicted in the beginning.
It is worth noting that this should be considered as the closest point to which the foot should be placed to restore from the push. After the impact, the momentum based controller is responsible of keeping the robot in an erect position on two feet. 

Summarizing our approach, in $t = 0$ we compute the desired foot location $x^r_\text{foot} \in \mathbb{R}^3$, corresponding to the position the Capture Point will have at $t = t_\text{step}$.
At every controller period, i.e. every $10 \mathrm{ms}$ in our implementation, we refine the previously computed quantity by reading the actual state of the robot.
This allows us to increase the robustness of the strategy to modeling errors.

\subsection{Reference Generation}
Most of the Capture Point-based implementations on real robots generate reference joint trajectories which are then tracked by the position controller.
While this approach usually results in well-tracked trajectories, it is usually not sufficiently robust to environment discrepancies or additional external disturbances.
Our recovery strategy, instead of applying the obtained control actions directly to the robot by means of position control, follows a different route.
As Figure \ref{fig:strategy} shows, we choose to feed the generated actions to the momentum-based whole-body torque controller briefly described in Section \ref{sec:momentum}.
It is thus necessary to coordinate the actions chosen by the step strategy with the ones taken by the low level balancing controller, which actually commands the robot.

To accomplish the aforementioned coordination objective we feed the momentum-based controller presented in Section \ref{sec:momentum} with a linear momentum reference $H_\text{lin}^r, \dot{H}_\text{lin}^r$ and with a reference for the joint configuration $q_j^r$.

The joint configuration $q_j^r$ is obtained starting from the desired foot position as predicted by the Capture Point criterion and described in the previous section. The desired foot orientation can be chosen at will, therefore we decide to keep it parallel to the stance foot for the sake of simplicity. The foot pose is then transformed into desired joint positions by inverse kinematics \cite{Pattacini2010}. 
This algorithm handles the redundancies by considering joint limits and picking the solution closest to current robot configuration.
Notice that joint references may not be tracked because of the prioritized structure of the momentum controller presented in Section \ref{sec:momentum}. The center of mass trajectory, therefore, should be ``compatible'' with the desired joint configurations. For example, this trajectory may be the generated through forward kinematics given the computed $q_j^r$ opportunely smoothed \cite{Pattacini2010} to avoid being heavily influenced by a moving foot reference.
The desired linear momentum is then obtained by setting $H^r_{\text{lin}} = m\dot{x}_\text{CoM}^r.$
The correct definition of this trajectory is necessary to perform the step movement correctly. Indeed, the lack of ``coordination'' between the two controller tasks may lead to an erratic motion, causing the robot to fall. Finally, the adoption of a torque controller may allow to recover balance even if the foot positioning is imperfect, as the intrinsic compliance helps adapting to the ground surface.
%

\section{Simulations and experimental results}
\label{sec:experiments}

This section presents the experimental results performed both in simulation and on the real iCub humanoid robot, that for the purpose of the proposed recovery strategy, is endowed with $23$ degrees of freedom.
In both experiments the robot is controlled by the momentum-based torque controller described in Section \ref{sec:momentum}.
The robot starts the experiment balancing on two feet and then moves on one foot balancing.
At this point, the robot is pushed.


\subsection{Experiments in Gazebo simulator}

\begin{figure}[t]
  \centering
    \includegraphics[width=.8\columnwidth]{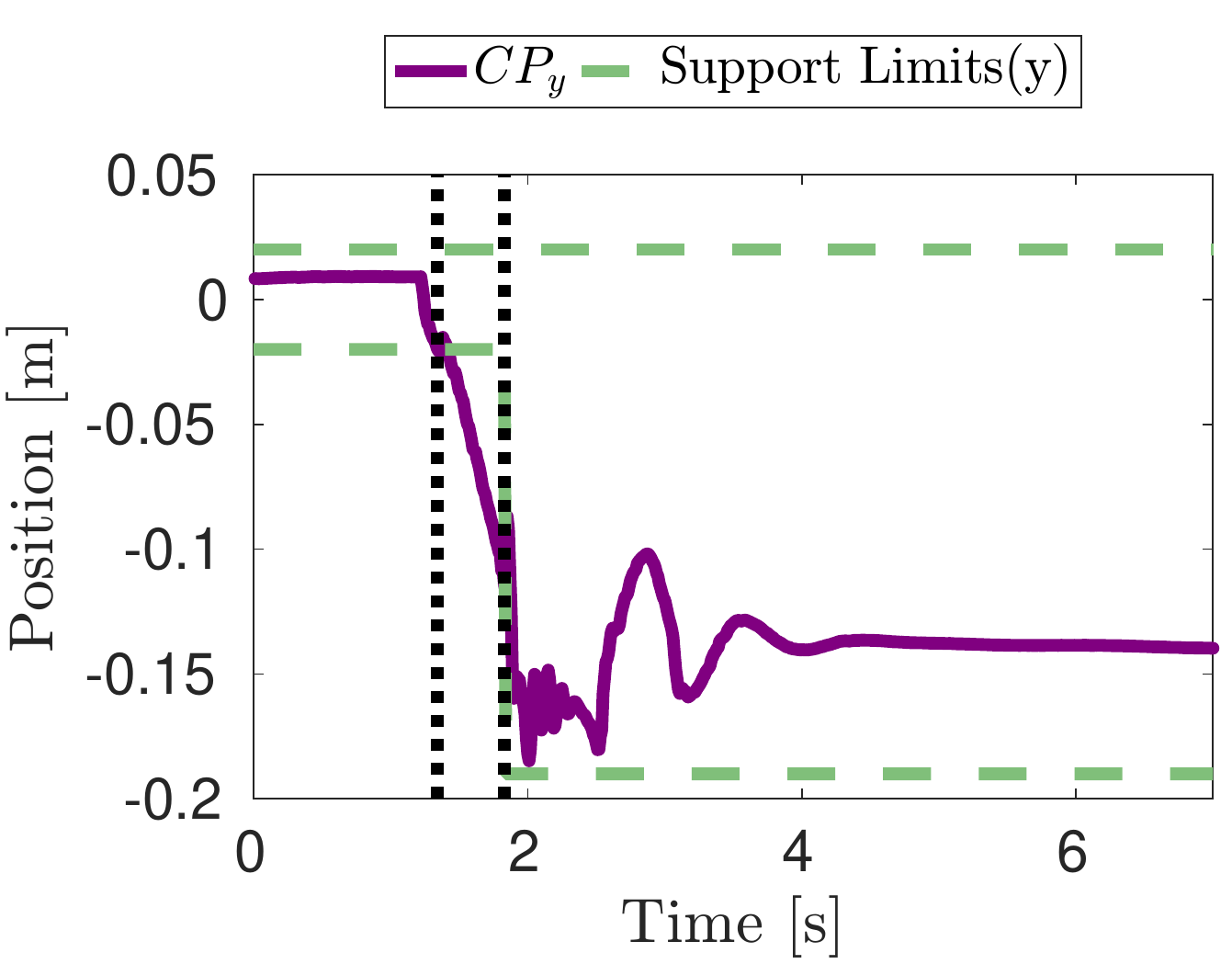}
  \caption{Evolution of the Capture Point $y-$component (the push is lateral) during the simulation experiment. 
  The first vertical dotted line denotes the push instant. The second one indicates when the right foot hits the ground. The horizontal dotted lines in green represent an approximation of the support convex hull during the step. }
  \label{fig:icp}
\end{figure}
%
\begin{figure*}[t]
	\centering
	\includegraphics[width=.8\columnwidth]{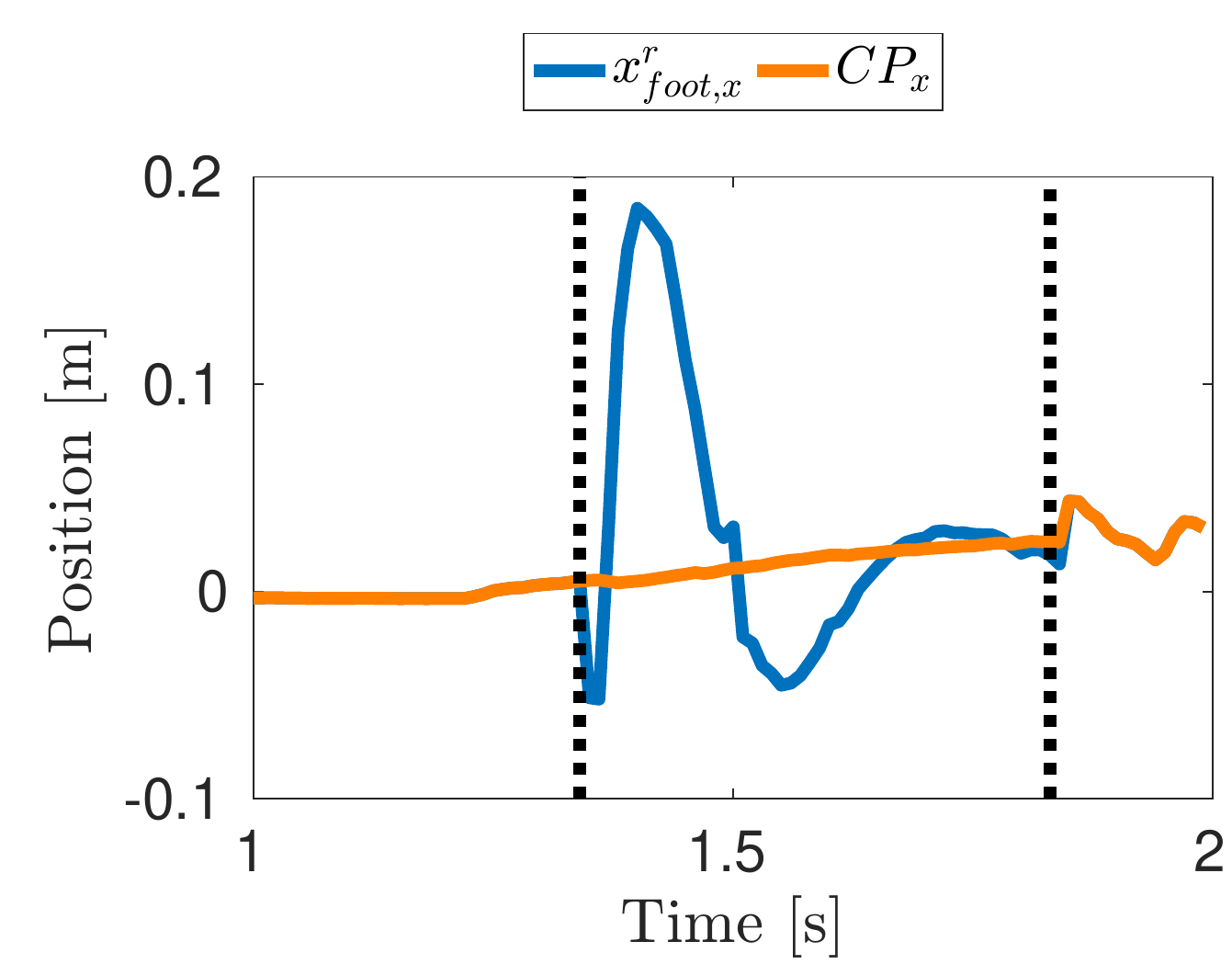}
	\qquad\qquad\qquad
	\includegraphics[width=.8\columnwidth]{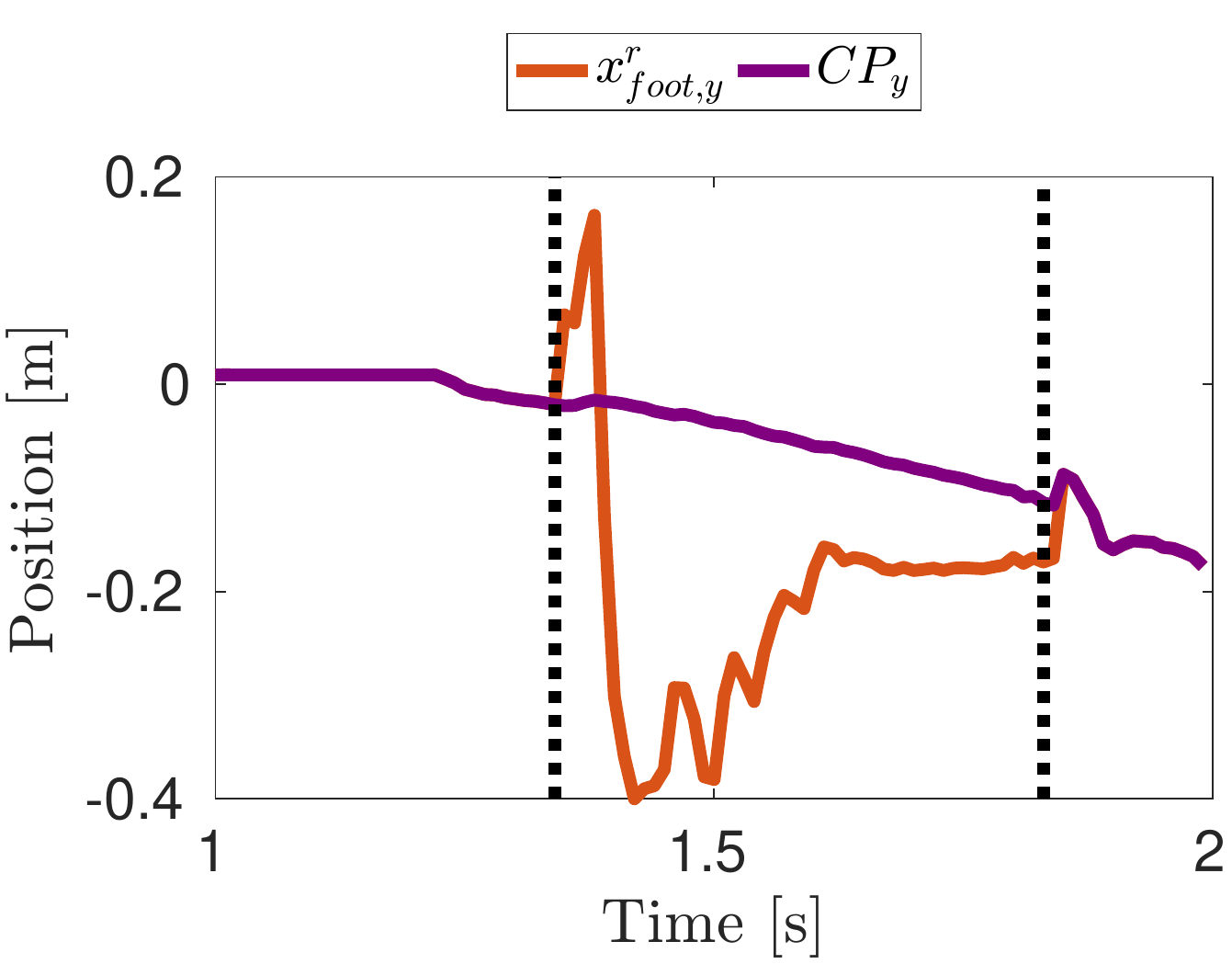}
	\caption{Preview capability of Eq. \eqref{eq:icp_estimation}. Each point of the blue and the red line indicates the estimation of the $x$ and $y$ coordinates of the Capture Point at the impact. The two dotted lines denotes the interval in which the step is taken.}
	\label{fig:cpVSicp}
\end{figure*}
\begin{figure}[t]
	\centering
	\includegraphics[width=.8\columnwidth]{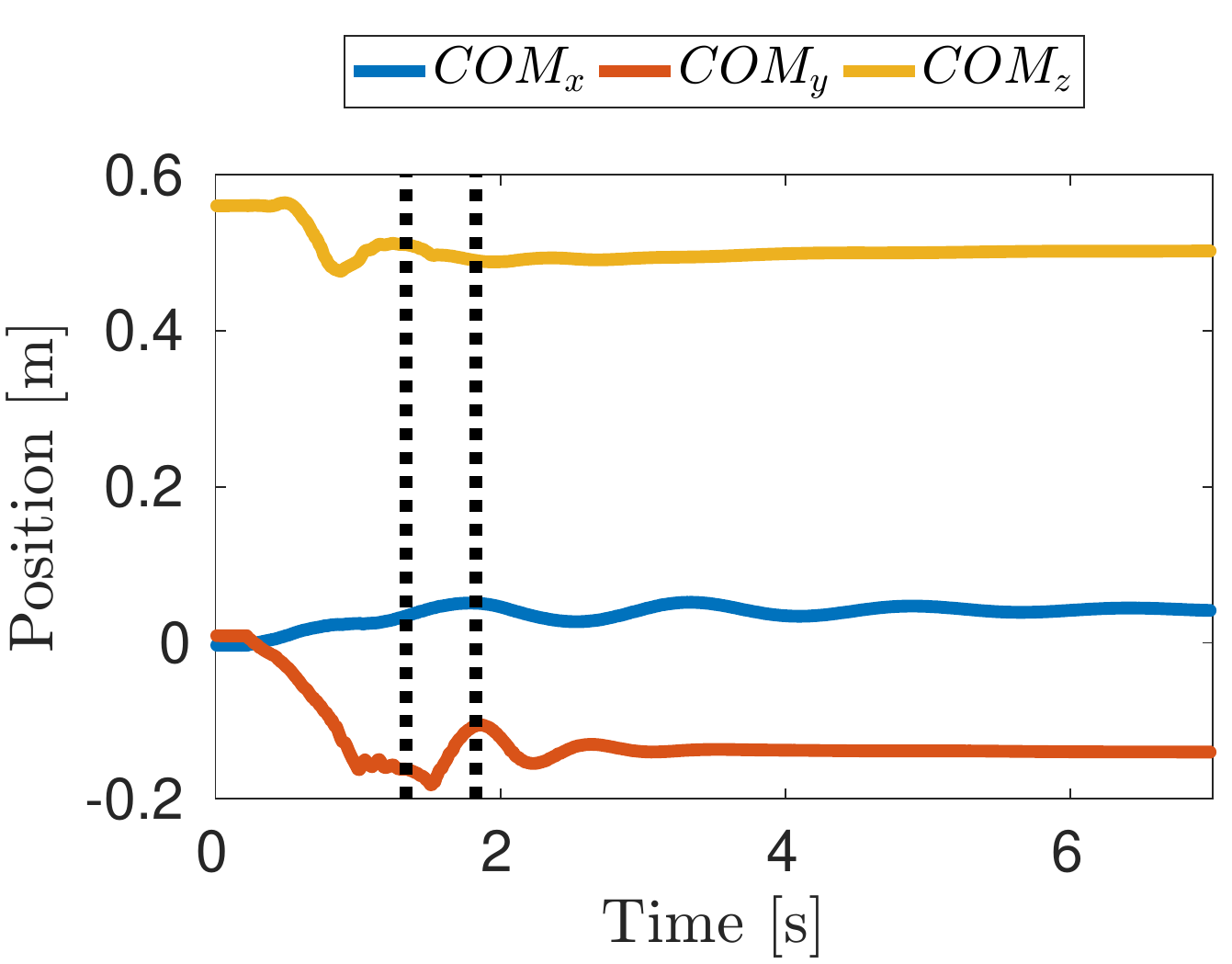}
	\caption{Evolution of the center of mass during the simulation experiment. 
		The vertical dotted lines have the same meaning of Figure \ref{fig:icp}.
		After the second push, the reference for the center of mass is moved to the middle of the two feet.}
	\label{fig:com}
\end{figure}

We first test the proposed step strategy in simulation using the Gazebo simulator \cite{Koenig04} and the corresponding Yarp plugins \cite{YarpGazebo2014}. 

A push is applied to the robot at $t = 1.35\mathrm{s}$.
Because the push is strong enough to bring the Capture Point out of the foot, as it can be seen in Figure \ref{fig:icp}, the condition described in Section \ref{sec:trigger} triggers. 
As a consequence, by using the FLIP model (see Section \ref{sec:foot_placement}), we generate the reference for the position of the right foot, according to Eq. \eqref{eq:icp_estimation}.
Figure \ref{fig:cpVSicp} shows the behaviour of the Capture Point and the prediction given by Eq. \eqref{eq:icp_estimation} during the step, which will be used as a reference for the foot placement. It is worth noting that, close to the end of the step, the desired footstep reaches the Capture Point. 
The CoM position is represented in Figure \ref{fig:com}, proving the capability of the robot to recover from the push.


\subsection{Experiments on the iCub robot}
\begin{figure}[t]
	\centering
    \includegraphics[width=.8\columnwidth]{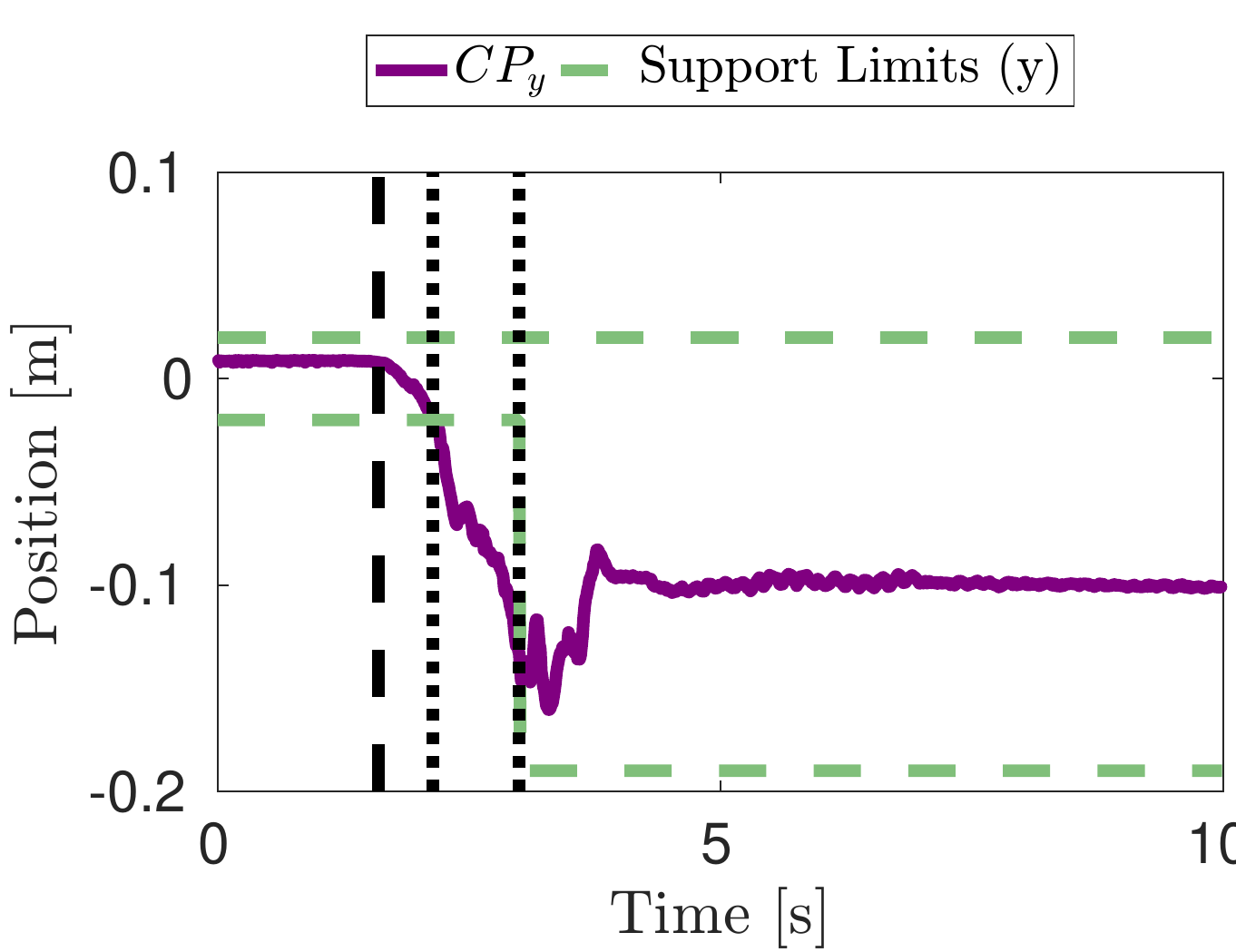}
	\caption{Evolution of the Capture Point $y-$component (the push is lateral) during the experiment on the robot. 
		The first bold line denotes the occurrence of the push which leads the CP to exit the support polygon at $t = 2.1\mathrm{s}$. From here the step is taken approximately up to $t = 3\mathrm{s}$.}
	\label{fig:icp_rb}
\end{figure}

\begin{figure}[t]
	\centering
	\includegraphics[width=.8\columnwidth]{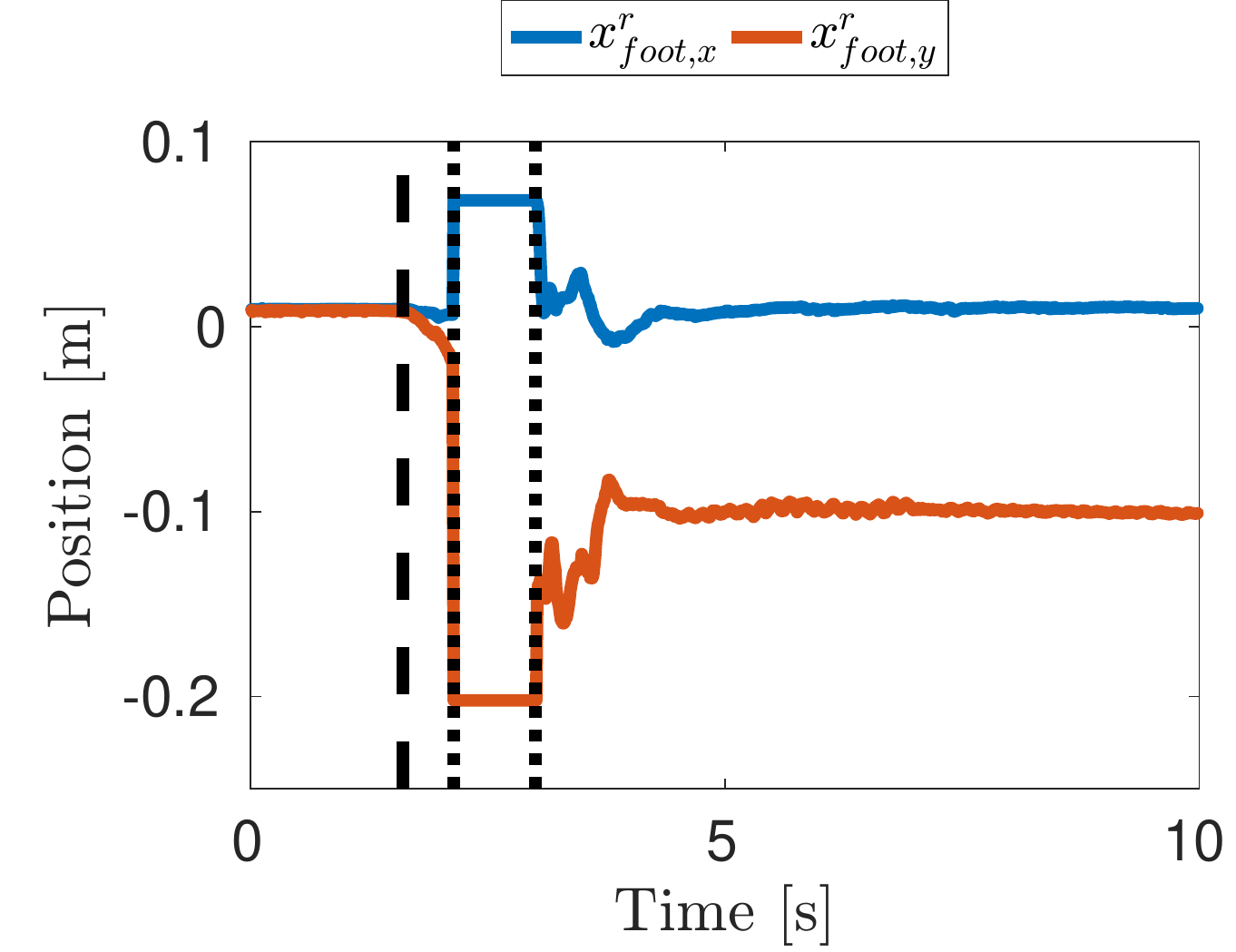}
	\caption{Right foot cartesian reference during the experiment on the real robot. When the robot is again in double support, it coincides with the Capture Point and the step is no more triggered. The vertical dotted lines have the same meaning of Figure \ref{fig:icp_rb}}
	\label{fig:cp_rb}
\end{figure}

\begin{figure}[t]
	\centering
	\includegraphics[width=.8\columnwidth]{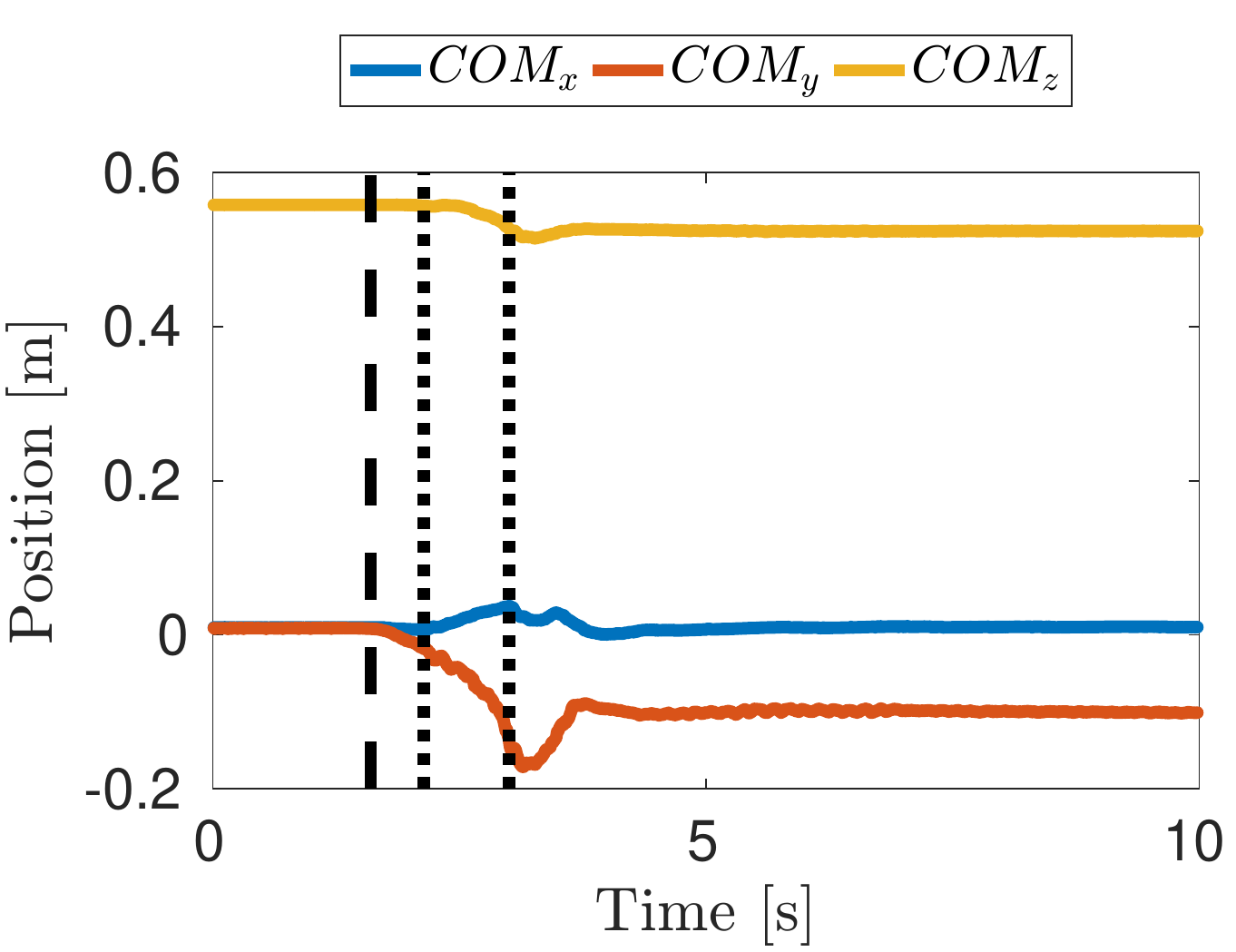}
	\caption{Robot center of mass position throughout the whole step, from the push to the full recovery. The first two vertical lines indicate when the push starts and when the step trigger occurs. The third line is the impact.}
	\label{fig:com_rb}
\end{figure}


The presented step recovery strategy has been implemented also on the iCub humanoid robot.
The robot has been pushed at $t = 1.6\mathrm{s}$, and, as a result, the Capture Point exits the support polygon at time $t = 2.1\mathrm{s}$, as it can be seen from Figure \ref{fig:icp_rb}. 
Differently from the simulation experiments, the foot reference position has been obtained using the 3D linear inverted pendulum model.
This choice has been dictated by not precise measurements of the CoP.
Furthermore, this model usually generates larger step references with respect to the FLIP model, thus making it less likely to underestimate the foot step length.
An additional benefit of this simpler model is the constancy of the generated reference. 
Indeed, the predicted Capture Point at the push instant is given by Eq. \eqref{eq:capture_point_predicted_lipm}, which depends only on the initial conditions and on the constant stance foot position.
Because of the constancy of the reference, the inverse kinematics function is called only once.
Given that it usually requires about $30\mathrm{ms}$ to yield its results, the performance of the controller benefits from this fact.
The cartesian reference for the right foot is in Figure~\ref{fig:cp_rb}.

Figure \ref{fig:com_rb} shows the center of mass trajectory. 
After the robot returns to double support, which occurs at $t = 3 \mathrm{s}$, the center of mass is properly stabilized in the middle of the two feet, thus proving the effectiveness of the recovery strategy.


When the foot impacts the ground, the force on the foot presents a discontinuity with an amplitude comparable to the weight of the robot itself.
It is in this case that the torque control exhibits its advantages helping absorbing the impact through its intrinsic compliance, without the need to add it explicitly in the formulation.

%

\section{CONCLUSIONS}
\label{sec:conclusions}
In this paper we presented a control architecture to achieve push recovery under strong disturbances by using the concept of the Capture Point.
Differently from most of the state-of-the-art approaches, which control the robot in position, the obtained references are then fed to a momentum-based torque controller thus allowing us to retain the advantages of the torque control while interacting with the environment.
Simulation results and experiments on the iCub humanoid robots validate the soundness of the proposed approach.

As a future work, multiple and possibly crossing steps will be employed, in order to face an even wider set of perturbations.
In addition, as discussed in Section \ref{sec:momentum} the considered constrained dynamical system is a hybrid system as the dynamics changes depending on the contact configuration. 
A more thoughtful theoretical analysis about the mitigation of torque discontinuities during the state transitions should be performed. We intend to investigate this subject in the near future.
Another important element we will work on regards the model complexity.
Most of the current state-of-the-art approaches, including the Capture Point, relies on simplified and approximated models of the full free-floating dynamics of the robot.
While these approximations allow us to perform simpler mathematical analysis and obtain control actions without expensive computations, we think that to operate in highly dynamic environments the full whole-body model should be exploited.



\addcontentsline{toc}{section}{References}

\bibliography{IEEEabrv,Bibliography}

\begin{thebibliography}{10}
\providecommand{\url}[1]{#1}
\csname url@rmstyle\endcsname
\providecommand{\newblock}{\relax}
\providecommand{\bibinfo}[2]{#2}
\providecommand\BIBentrySTDinterwordspacing{\spaceskip=0pt\relax}
\providecommand\BIBentryALTinterwordstretchfactor{4}
\providecommand\BIBentryALTinterwordspacing{\spaceskip=\fontdimen2\font plus
\BIBentryALTinterwordstretchfactor\fontdimen3\font minus
  \fontdimen4\font\relax}
\providecommand\BIBforeignlanguage[2]{{%
\expandafter\ifx\csname l@#1\endcsname\relax
\typeout{** WARNING: IEEEtran.bst: No hyphenation pattern has been}%
\typeout{** loaded for the language `#1'. Using the pattern for}%
\typeout{** the default language instead.}%
\else
\language=\csname l@#1\endcsname
\fi
#2}}

\bibitem{Pratt2006}
J.~Pratt, J.~Carff, S.~Drakunov, and A.~Goswami, ``Capture {P}oint: {A} {S}tep
  toward {H}umanoid {P}ush {R}ecovery,'' in \emph{Humanoid Robots, 2006 6th
  IEEE-RAS International Conference on}, Dec 2006, pp. 200--207.

\bibitem{Kajita2001}
S.~Kajita, F.~Kanehiro, K.~Kaneko, K.~Yokoi, and H.~Hirukawa, ``The 3{D}
  {L}inear {I}nverted {P}endulum {M}ode: {A} simple modeling for a biped
  walking pattern generation,'' in \emph{Intelligent Robots and Systems, 2001.
  IEEE/RSJ International Conference on}, vol.~1, 2001.

\bibitem{park1998biped}
J.~H. Park and K.~D. Kim, ``Biped robot walking using gravity-compensated
  inverted pendulum mode and computed torque control,'' in \emph{Robotics and
  Automation, 1998. Proceedings. 1998 IEEE International Conference on},
  vol.~4.\hskip 1em plus 0.5em minus 0.4em\relax IEEE, 1998, pp. 3528--3533.

\bibitem{goswami2014direction}
A.~Goswami, S.-k. Yun, U.~Nagarajan, S.-H. Lee, K.~Yin, and S.~Kalyanakrishnan,
  ``Direction-changing fall control of {H}umanoid robots: theory and
  experiments,'' \emph{Autonomous Robots}, vol.~36, no.~3, 2014.

\bibitem{Frontiers2015}
F.~Nori, S.~Traversaro, J.~Eljaik, F.~Romano, A.~Del~Prete, and D.~Pucci,
  ``i{C}ub whole-body control through force regulation on rigid noncoplanar
  contacts,'' \emph{Frontiers in Robotics and AI}, vol.~2, no.~6, 2015.

\bibitem{Ott2011}
C.~Ott, M.~Roa, and G.~Hirzinger, ``Posture and {B}alance {C}ontrol for {B}iped
  {R}obots based on {C}ontact {F}orce {O}ptimization,'' in \emph{Humanoid
  Robots (Humanoids), 2011 11th IEEE-RAS International Conference}.

\bibitem{koolen2016design}
T.~Koolen, S.~Bertrand, G.~Thomas, T.~De~Boer, T.~Wu, J.~Smith, J.~Englsberger,
  and J.~Pratt, ``Design of a {M}omentum-{B}ased {C}ontrol {F}ramework and
  {A}pplication to the {H}umanoid {R}obot {A}tlas,'' \emph{International
  Journal of Humanoid Robotics}, vol.~13, no.~01, 2016.

\bibitem{stephens2007humanoid}
B.~Stephens, ``Humanoid {P}ush {R}ecovery,'' in \emph{Humanoid Robots, 2007 7th
  IEEE-RAS International Conference on}.\hskip 1em plus 0.5em minus 0.4em\relax
  IEEE, 2007.

\bibitem{morisawa2010combining}
M.~Morisawa, F.~Kanehiro, K.~Kaneko, N.~Mansard, J.~Sola, E.~Yoshida, K.~Yokoi,
  and J.-P. Laumond, ``Combining {S}uppression of the {D}isturbance and
  {R}eactive {S}tepping for {R}ecovering {B}alance,'' in \emph{Intelligent
  Robots and Systems (IROS), 2010 IEEE/RSJ International Conference on}.\hskip
  1em plus 0.5em minus 0.4em\relax IEEE, 2010, pp. 3150--3156.

\bibitem{stephens2010pushforce}
B.~J. Stephens and C.~G. Atkeson, ``Push {R}ecovery by {S}tepping for
  {H}umanoid {R}obots with {F}orce {C}ontrolled {J}oints,'' in \emph{Humanoid
  Robots (Humanoids), 2010 10th IEEE-RAS International Conference on}.\hskip
  1em plus 0.5em minus 0.4em\relax IEEE, 2010, pp. 52--59.

\bibitem{koolen2012capturability}
T.~Koolen, T.~De~Boer, J.~Rebula, A.~Goswami, and J.~Pratt,
  ``Capturability-{B}ased {A}nalysis and {C}ontrol of {L}egged {L}ocomotion,
  {P}art 1: {T}heory and {A}pplication to {T}hree {S}imple {G}ait {M}odels,''
  \emph{The International Journal of Robotics Research}, vol.~31, no.~9, 2012.

\bibitem{vukobratovic2004zero}
M.~Vukobratovi{\'c} and B.~Borovac, ``Zero-{M}oment {P}oint: thirty five years
  of its life,'' \emph{International Journal of Humanoid Robotics}, vol.~1,
  no.~01, pp. 157--173, 2004.

\bibitem{englsberger2011bipedal}
J.~Englsberger, C.~Ott, M.~A. Roa, A.~Albu-Sch{\"a}ffer, and G.~Hirzinger,
  ``Bipedal walking control based on {C}apture {P}oint dynamics,'' in
  \emph{2011 IEEE/RSJ International Conference on Intelligent Robots and
  Systems}.

\bibitem{morisawa2012balance}
M.~Morisawa, S.~Kajita, F.~Kanehiro, K.~Kaneko, K.~Miura, and K.~Yokoi,
  ``Balance {C}ontrol based on {C}apture {P}oint {E}rror {C}ompensation for
  {B}iped {W}alking on {U}neven {T}errain,'' in \emph{2012 12th IEEE-RAS
  International Conference on Humanoid Robots (Humanoids 2012)}.\hskip 1em plus
  0.5em minus 0.4em\relax IEEE, 2012, pp. 734--740.

\bibitem{krause2012stabilization}
M.~Krause, J.~Englsberger, P.-B. Wieber, and C.~Ott, ``Stabilization of the
  {C}apture {P}oint {D}ynamics for {B}ipedal {W}alking based on {M}odel
  {P}redictive {C}ontrol,'' \emph{IFAC Proceedings Volumes}, vol.~45, no.~22,
  2012.

\bibitem{ramos2014whole}
O.~E. Ramos, N.~Mansard, and P.~Soueres, ``Whole-body {M}otion {I}ntegrating
  the {C}apture {P}oint in the {O}perational {S}pace {I}nverse {D}ynamics
  {C}ontrol,'' in \emph{2014 IEEE-RAS International Conference on Humanoid
  Robots}.\hskip 1em plus 0.5em minus 0.4em\relax IEEE, 2014, pp. 707--712.

\bibitem{pratt2012capturability}
J.~Pratt, T.~Koolen, T.~De~Boer, J.~Rebula, S.~Cotton, J.~Carff, M.~Johnson,
  and P.~Neuhaus, ``Capturability-{B}ased {A}nalysis and {C}ontrol of {L}egged
  {L}ocomotion, {P}art 2: {A}pplication to {M2V2}, a {L}ower {B}ody
  {H}umanoid,'' \emph{The International Journal of Robotics Research}, 2012.

\bibitem{metta2005robotcub}
G.~Metta, G.~Sandini, D.~Vernon, D.~Caldwell, N.~Tsagarakis, R.~Beira,
  J.~Santos-Victor, A.~Ijspeert, L.~Righetti, G.~Cappiello, \emph{et~al.},
  ``The {R}obot{C}ub project-an open framework for research in embodied
  cognition,'' \emph{Humanoids Workshop, IEEE--RAS International Conference on
  Humanoid Robots, December}, 2005.

\bibitem{nava16}
G.~Nava, F.~Romano, F.~Nori, and D.~Pucci, ``Stability {A}nalysis and {D}esign
  of {M}omentum-based {C}ontrollers for {H}umanoid {R}obots,''
  \emph{Intelligent Robots and Systems (IROS) 2016. IEEE International
  Conference on}, 2016.

\bibitem{sardain2004forces}
P.~Sardain and G.~Bessonnet, ``Forces {A}cting on a {B}iped {R}obot. {C}enter
  of {P}ressure-{Z}ero {M}oment {P}oint,'' \emph{Systems, Man and Cybernetics,
  Part A: Systems and Humans, IEEE Transactions on}, vol.~34, no.~5, 2004.

\bibitem{Pattacini2010}
U.~Pattacini, F.~Nori, L.~Natale, G.~Metta, and G.~Sandini, ``{An Experimental
  Evaluation of a Novel Minimum-Jerk Cartesian Controller for Humanoid
  Robots},'' in \emph{Intelligent Robots and Systems (IROS), IEEE/RSJ
  International Conference on}.\hskip 1em plus 0.5em minus 0.4em\relax IEEE,
  2010, pp. 1668--1674.

\bibitem{Koenig04}
N.~Koenig and A.~Howard, ``Design and {U}se {P}aradigms for {G}azebo, an
  {O}pen-{S}ource {M}ulti-{R}obot {S}imulator,'' \emph{Intelligent Robots and
  Systems, 2004. (IROS 2004) 2004 IEEE/RSJ International Conference on}, 2004.

\bibitem{YarpGazebo2014}
E.~Mingo, S.~Traversaro, A.~Rocchi, M.~Ferrati, A.~Settimi, F.~Romano,
  L.~Natale, A.~Bicchi, F.~Nori, and N.~G. Tsagarakis, ``{Y}arp {B}ased
  {P}lugins for {G}azebo {S}imulator,'' in \emph{2014 Modelling and Simulation
  for Autonomous Systems Workshop (MESAS)}, 2014.

\end{thebibliography}

\end{document}